\documentclass[10pt,twocolumn,letterpaper]{article}
\usepackage[pagenumbers]{wacv} 
\PassOptionsToPackage{numbers, compress, sort}{natbib}
\usepackage{natbib}
\usepackage{graphicx}
\usepackage{booktabs}
\usepackage{subcaption}
\usepackage{xcolor}
\usepackage{arydshln}
\usepackage{pifont}
\usepackage{microtype}
\usepackage{multirow}
\newcommand{\OURMODEL}{GaB}
\newcommand{\OURMODELCLA}{GaB-classifier}
\newcommand{\OURMODELCLU}{GaB-clustering}

\newcommand{\range}[2]{#1\mathinner {\ldotp \ldotp}#2}
\newcommand{\twodots}{{\ldotp \ldotp}}

\definecolor{aqua}{rgb}{0.0, 1.0, 1.0}
\definecolor{ForestGreen}{RGB}{34,139,34}
\definecolor{CoreCircuits}{RGB}{123,241,168}
\definecolor{CoreHam}{RGB}{254,200,154}
\definecolor{CoreInat}{RGB}{189,224,254}
\newcommand{\cmark}{\textcolor{ForestGreen}{\ding{51}}}
\newcommand{\xmark}{\textcolor{red}{\ding{55}}}

\usepackage{graphicx}
\usepackage{amsmath}
\usepackage{amssymb}
\usepackage{booktabs}
\usepackage{mathtools}
\DeclarePairedDelimiter\ceil{\lceil}{\rceil}

\usepackage[pagebackref,breaklinks,colorlinks]{hyperref}

\usepackage[capitalize]{cleveref}
\crefname{section}{Sec.}{Secs.}
\Crefname{section}{Section}{Sections}
\Crefname{table}{Table}{Tables}
\crefname{table}{Tab.}{Tabs.}

\def\wacvPaperID{1480} 
\def\confName{WACV}
\def\confYear{2025}

\begin{document}

\title{One VLM to Keep it Learning: Generation and Balancing for Data-free Continual Visual Question Answering}

\author{Deepayan Das\textsuperscript{$1$} \quad
Davide Talon\textsuperscript{$2$} \quad
Massimiliano Mancini\textsuperscript{$1$} \quad
\\
Yiming Wang\textsuperscript{$2$} \quad
Elisa Ricci\textsuperscript{$1,2$} \\
\small
$^1$University of Trento \quad \quad $^2$Fondazione Bruno Kessler \\
\small
\texttt{\{deepayan.das, massimiliano.mancini, e.ricci\}@unitn.it}\\
\small
\texttt{\{dtalon, ywang\}@fbk.eu}\\
}

\maketitle

\begin{abstract}
   Vision-Language Models (VLMs) have shown significant promise in Visual Question Answering (VQA) tasks by leveraging web-scale multimodal datasets. However, these models often struggle with continual learning due to catastrophic forgetting when adapting to new tasks. As an effective remedy to mitigate catastrophic forgetting, rehearsal strategy uses the data of past tasks upon learning new task. However, such strategy incurs the need of storing past data, which might not be feasible due to hardware constraints or privacy concerns. In this work, we propose the first data-free method that leverages the language generation capability of a VLM, instead of relying on external models, to produce pseudo-rehearsal data for addressing continual VQA. Our proposal, named as GaB, generates pseudo-rehearsal data by posing previous task questions on new task data. Yet, despite being effective, the distribution of generated questions skews towards the most frequently posed questions due to the limited and task-specific training data. To mitigate this issue, we introduce a pseudo-rehearsal balancing module that aligns the generated data towards the ground-truth data distribution using either the question meta-statistics or an unsupervised clustering method. We evaluate our proposed method on two recent benchmarks, \ie VQACL-VQAv2 and CLOVE-function benchmarks. GaB outperforms all the data-free baselines with substantial improvement in maintaining VQA performance across evolving tasks, while being on-par with methods with access to the past data. Code and models are available at \url{https://github.com/Deepayan137/GaB}. 
\end{abstract}


\section{Introduction}
Visual Question Answering (VQA) has been a prominent research challenge since its introduction a decade ago~\citep{antol2015vqa}, aiming to mimic the real-world scenarios in which an agent can interpret and reason about the visual content, \eg~the object attributes, their relations, and commonsense~\citep{goyal2017making, hudson2019gqa, smith2023construct, zhang2023vqacl} through natural language questions.
One of the main challenges of VQA is to bridge the multimodal gap as both visual and textual understanding are required, and their knowledge representation should be integrated. 
Recent Vision Language Models (VLMs), \eg, BLIP-2~\cite{li2023blip} and LLaVa~\cite{liu2024visual}, have greatly enhanced such multimodal understanding thanks to their web-scale pre-training with image-text pairs. 
\begin{figure}[t!]
    \centering
    \includegraphics[width=1\linewidth]{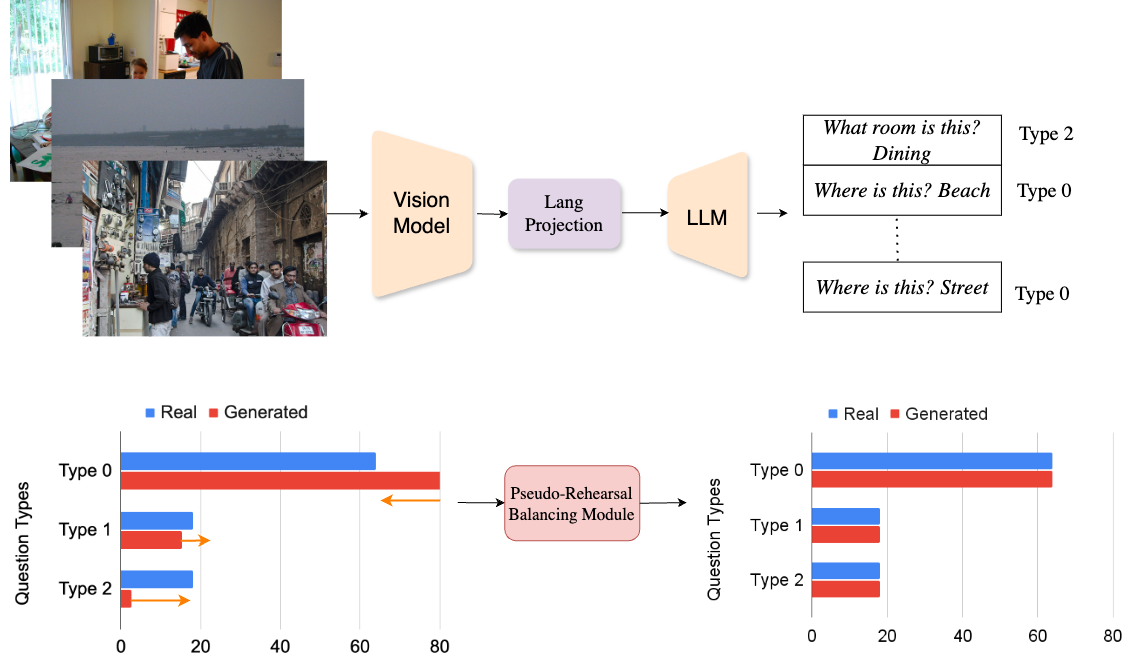}
    \caption{Key ideas. \textbf{Top:} we explore the language generation capability of VLMs to synthesize pseudo-rehearsal data of previous tasks to mitigate catastrophic forgetting in continual VQA. \textbf{Bottom:} as pseudo-rehearsal data tends to skew towards particular question types, we further propose a pseudo-rehearsal balancing module to align such skewed distribution towards the ground-truth meta-statistics, effectively improving the task performance while avoiding forgetting.}
    \label{fig:teaser}
\end{figure}
VLMs in addressing the VQA task, have demonstrated a good generalization ability, yet they require continuous model updates to cope with new knowledge and to handle new types of reasoning skills~\citep{smith2023construct, yu2024boosting}.
The main challenge in updating VQA models in such continual learning setup, a.k.a.~VQACL~\cite{zhang2023vqacl}, is to avoid the famous \textit{catastrophic forgetting} problem, \ie, without hindering the performance of previously learned tasks upon learning new ones~\citep{wang2024comprehensive, smith2023construct, zhang2023vqacl, greco2019psycholinguistics, hayes2020remind}. 
One common remedy to catastrophic forgetting is via data rehearsal, i.e., replay of the data samples of prior tasks during the model updates for new tasks~\citep{rebuffi2017icarl, chaudhry2019tiny, buzzega2020dark}. However, such prior data might not be available due to privacy or proprietary concerns~\cite{voigt2017eu}. 
\textit{Can we address VQACL without the access to data samples of previous tasks?} Prior methods leverage pseudo-rehearsal by generating data samples of prior task~\citep{lei2023symbolic, smith2023construct, shin2017continual}. 
Such pseudo-rehearsal methods are capable of mitigating catastrophic forgetting in a data-free manner. 
However, existing methods leverage additional models to produce rehearsal samples~\citep{lei2023symbolic,shin2017continual} or computationally demanding adversarial samples~\citep{smith2023construct}. 

\textit{In the era of VLMs, can we exploit their generative language model to generate data samples?} In this work, we explore this direction, instead of relying on additional models. We propose \textbf{\OURMODEL}, the first data-free method that leverages the language module of a VLM to address VQACL by generating question-answer pairs to mimic old tasks. 
Specifically, we generate questions and pseudo-answers about previous tasks by conditioning on the current visual data at hand. Crucially, the lack of diversity of questions related to the same task leads to generation collapse to data frequently occurring questions, inhibiting the VQA model capabilities (as shown in Fig.~\ref{fig:teaser}).
To mitigate this problem, we balance the generated multimodal data: we initially learn to categorize the questions that are posed during learning new tasks and later categorize generated data accordingly. Hence, the rehearsal buffer is re-sampled ensuring that training data and generated question type distributions align.

We evaluate \OURMODEL{} on continual learning VQA benchmarks, namely VQACL-VQAv2~\citep{zhang2023vqacl} and CLOVE-function~\citep{lei2023symbolic} and compare the proposed approach with respect to state-of-the-art and continual learning baselines. Results show a large margin improvement on all considered settings among data-free methods. 

To summarize, our contributions are three-fold:
\begin{itemize}
    \item We propose the first data-free method for VQACL, leveraging the language model inherent in VLMs to generate pseudo-rehearsal data, achieving the best performance among data-free methods on VQACL-VQAv2~\citep{zhang2023vqacl} and CLOVE-function~\citep{lei2023symbolic}.
    \item We propose to jointly learn a task-specific projection module for generating question-answer pairs conditioned on images of the current task, to then synthesize data for the the rehearsal buffer.
    \item We propose a novel and effective pseudo-rehearsal balancing module to mitigate the skewed distribution of the VLM-generated question types.
\end{itemize}

\begin{figure*}[ht]
    \centering
    \includegraphics[width=0.8\linewidth]{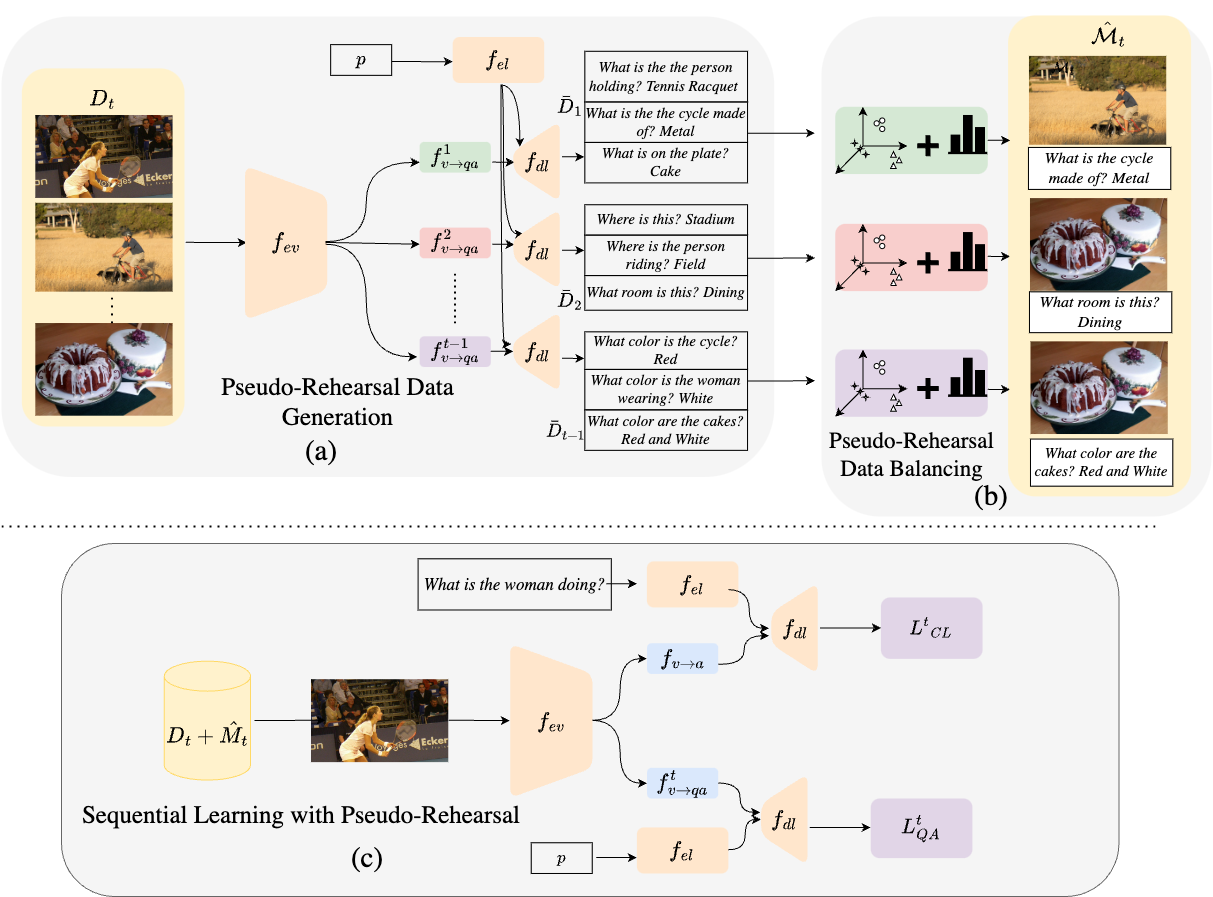}
    \caption{\textbf{Architecture of the proposed data-free method \OURMODEL{} for addressing VQACL.}
    \textbf{(a)} At task $t$, past task-specific projection heads $f^s_{v\to qa}, s=\range{1}{t-1}$ are used to generate pseudo-rehearsal data with question-answer pairs about old tasks on current task images. \textbf{(b)} Pseudo-rehearsal samples undergo a balancing process through a module designed to ensure that under-represented question types are adequately represented. Finally \textbf{(c)} uses the pseudo-rehearsal data to mitigate the forgetting in the sequential learning scenario and data $D_t$ to learn the current question-answer projector $f^t_{v \to qa}$.
}
\label{fig:approach}
\end{figure*}
\section{Related Work}
In the following, we review related work on continual learning and visual question answering.
\paragraph{Continual Learning.} Continual learning (CL) addresses the problem of learning new knowledge while avoiding forgetting old concepts, a problem commonly known as catastrophic forgetting~\citep{kirkpatrick2017overcoming, french1999catastrophic}.  
Depending on the strategy to mitigate forgetting, CL approaches can be divided into different categories. 
Regularization-based methods incorporate additional regularization terms into the objective functions when learning a new task~\citep{jung2016less, kirkpatrick2017overcoming, li2017learning, zenke2017continual, aljundi2018memory, schwarz2018progress}. 
Instead, rehearsal-based techniques replay previous tasks samples alongside new task data~\citep{rebuffi2017icarl, chaudhry2019tiny, buzzega2020dark, tiwari2022gradient, tang2022learning}. However, storing data may be limiting in settings with privacy or copyright concerns. 
Architecture-based methods \citep{aljundi2017expert, li2019learn, rusu2016progressive, yoon2018lifelong, wang2022learning, wang2022dualprompt, krishna2019information, roy2023exemplar} allocate different model parameters for different tasks. 
Here we follow the second family of approaches, addressing forgetting by replaying knowledge. However, we exploit the capabilities of VLMs to generate pseudo question-answer pairs of previous tasks.
Contrary to previous work, we employ a data-free rehearsal strategy that do not revert the past task images~\citep{choi2021dual, yin2020dreaming, smith2021always} but mimic the old language question-answer pairs on incoming new visual samples.
\paragraph{Continual Learning and VLMs.} Vision Language foundation models~\citep{radford2021learning, li2023blip, alayrac2022flamingo} provide general knowledge that can serve as the basis for specialized models. By building on the features transferability and the initial off-the-shelf strong performance, recent work has investigated the adaptation of VLMs for the continual setting~\citep{zhang2023overcoming, singh2024controlling, yu2024boosting, xu2024advancing, smith2023construct}. While most of the work has focused on maintaining the zero-shot generalization capabilities of foundation VLMs~\citep{yu2024boosting, xu2024advancing}, multimodal continual learning approaches have relied on model expansion~\citep{zhou2023learning, wang2022s}, caption and label matching~\citep{cao2024generative}, uncertainty~\citep{jha2024clap4clip} and prompting~\citep{wang2023attriclip}. In contrast to \citep{kim2024vlm} building on VLMs for pseudo-labels refinement, our approach leverages the textual generation of question-answer pairs on new data. After the initial expansion of reply heads, the model replays previous questions, avoiding estimating the task identity at inference time~\cite{zhou2023learning, wang2022s}. 

\paragraph{Continual Learning and VQA.} Despite recent advancements in the classic VQA setting~\citep{li2023blip, dai2024instructblip, alayrac2022flamingo, liu2024visual}, limited work concentrates on the multi-modal continual learning VQA setting where a model is sequentially trained to address different skills~\citep{hayes2020remind, srinivasan2022climb, lei2023symbolic, smith2023construct, qian2023decouple, zhang2023vqacl, zheng2024beyond, nikandrou2022task, zhang2022cl}. VQACL~\citep{zhang2023vqacl} leverages previous tasks samples and the question-type, question-object separation. Instead, REMIND~\citep{hayes2020remind} builds on latent replay samples as a form of compressed training data storage. Thanks to scene-graphs provided from a pre-trained model, ~\citep{lei2023symbolic} maintains a symbolic replay memory collecting scene-graph and question-answer pairs. We set apart from replay-based approaches~\citep{hayes2020remind, lei2023symbolic, zhang2022cl} that collect statistics or a coreset of previous tasks samples. Similarly to~\citep{smith2023construct} we build on incoming data to generate pseudo-labeled replay samples, however, by building on VLM language generation capabilities to replicate previous tasks questions in a self-training manner, we avoid the large number of expensive adversarial negative samples.
While automatic question generation has been explored for question-answer rehearsal~\cite{sun2019lamol, krishna2019information}, previous work costly generate context for previous data samples~\cite{sun2019lamol} or requires past rehearsal data to access conditioning answer types~\cite{krishna2019information}. On the contrary, our pseudo-samples strategy leverage current task visual data as augmentation strategy to replay previous tasks.

\section{Method} \label{sec:method}
In this section we present~\OURMODEL{}, our novel data-free approach for continual VQA. We start with a formal definition of the VQACL task with the necessary notations. We then provide an overview of the proposed method, followed by a detailed description on each key element, including the \textit{pseudo-rehearsal data generation}, the \textit{pseudo-rehearsal data balancing}, and the \textit{sequential learning with pseudo-rehearsal}.
\paragraph{Problem formulation.} Continual Learning for Visual Question Answering is generally defined as learning on a sequence of $T$ VQA tasks querying about different set of properties. Let $D_t, t=\range{1}{T}$, be the only dataset available at task $t$. Each dataset is defined as $D_t = \{d_t^1, d_t^2, \twodots, d_t^{N_t}\}$, where \(d_t^i\) consists of an image-question-answer triplet $(x_t^i, q_t^i, a_t^i)$ and $N_t$ is the number of samples of the $t$-th task. Here, \(x_t^i\) denotes the image, $q_t^i$ the associated question, and $a_t^i$ is the correct answer. The goal of the VQA model $f$ is to correctly predict the answer $a_t^i$ from the associated visual input $x_t^i$ with question $q_t^i$, while mitigating the catastrophic forgetting of old tasks $s$, $s<t$.
In the following, when the task or the sample are clear from the context, we simplify the notation by dropping the associated indexes, e.g., $x, q, a$. 
\paragraph{Addressing VQACL with data rehearsal.} A common and effective strategy for addressing catastrophic forgetting is rehearsal~\cite{rebuffi2017icarl, chaudhry2019tiny, buzzega2020dark}. Rehearsal-based CL follows a simple idea: a buffer stores samples of old tasks $s<t$. This buffer is then merged with the current task data $D_t$ and used to train the model, reducing forgetting by replaying old knowledge.

Let $\mathcal{M}_{t}$ be a memory buffer of size $M$, containing data for tasks $s=\range{1}{t-1}$. Without loss of generality, we assume that $\mathcal{M}_t$ is balanced across tasks, i.e., it contains the same amount of samples from each seen dataset $\range{D_1}{D_{t-1}}$. 
A model can achieve high performance on the VQA task via fine-tuning and, depending on the content of $\mathcal{M}_t$, avoid forgetting. Once trained for $t$, the rehearsal strategy updates $\mathcal{M}_t$ to build the buffer for the next learning task $t+1$.
\paragraph{\OURMODEL{} Overview.}
As rehearsal-based methods incur the need for accessing real data in $\mathcal{M}_t$, which might not be feasible due to \eg, security or privacy constraints, we propose a data-free rehearsal strategy. Our method \OURMODEL{}, as shown in~Fig.~\ref{fig:approach}, leverages the language-generation capabilities within a VLM to synthesize pseudo-rehearsal data, \ie, question-answer pairs from prior tasks on current task images.  
Specifically, we consider $f$ as a VLM which is composed of multiple components: an image encoder $f_{ev}$ and a text encoder $f_{el}$ which project images and text into their respective embedding spaces; A projector $f_{v\to l}$ projects visual features in the language embedding space; Finally, a language decoder $f_{dl}$ decodes the embeddings into natural language answers $\hat{a}$. 
With $f_{dl}$, VLMs have the inherent ability of generating text given a visual input and a textual prompt. This property enables us to build a data-free rehearsal buffer, without storing any data of old tasks.

At task $t$, \OURMODEL{} on one side of the model sequentially learns the task $t$-specific projector, with the goal to reconstruct $D_t$ question-answer pairs. On the other, old tasks projectors are employed for the generation of previous question-answer pairs given the current task images (Section~\ref{sec:qa-generation}). 
To align the distribution of generated and real data, we employ a re-sampling strategy (Section~\ref{sec:generation-collapse}) that leverages the data question type meta-statistics to re-sample pseudo-questions and obtain an aligned replay buffer distribution. Finally, the model leverages the available pseudo-rehearsal data for sequential learning (Section~\ref{sec:sequential-learning}). Notice that \OURMODEL{} requires task IDs only during training to manage independent and task-specific generation of rehearsal data. A single, shared visual question-answering head is trained using a replay strategy, facilitating backward transfer by integrating knowledge from new tasks, with no task IDs.
\subsection{Pseudo-rehearsal data generation}
\label{sec:qa-generation}
In our continual learning framework, we employ the same VLM model trained for VQA to generate the pseudo-rehearsal data for mitigating the catastrophic forgetting problem. Similarly to the continual learning for the VQA head, we learn to pose questions and associated answers, by reconstructing the textual information of the data samples.
\paragraph{Learning to generate question-answers.} At task $t$ we train a task-specific projection module $f^t_{v\to qa}$ to generate question answer pairs $(q_t^i, a_t^i)$ given input image $x_t^i$. Specifically, let $f^t_{qa}$ represent the use of projector $f^t_{v\to qa}$ within $f$ and operator $\circ$ denote the language concatenation. We can define the question-answer generation objective as:
\begin{equation}
    \label{eq:qa-objective}
    \mathcal{L}^t_{QA} = \frac{1}{N_t} \sum_{(x,a,q)\in D_t} \mathcal{L}_{\text{lm}}(f^t_{qa}(x,p), q \circ a),
\end{equation}
where $p$ is a prompt used to query the model to generate question-answer pairs and the language modelling loss is defined as:
\begin{equation}
\mathcal{L}_{\text{lm}}(\hat{y}, y)=\text{CE}(\hat{y}, y),
\end{equation}
evaluating the token-level cross-entropy loss between predicted text $\hat{y}$ and ground truth one $y$. With Eq.~\eqref{eq:qa-objective} we obtain a projection module that allows us to generate question and answer pairs for task $t$ given \textit{any} arbitrary image $x$. {Previously learned ones $f^{s}_{v\to qa}$ are kept frozen, $s=\range{1}{t-1}$.} 
\paragraph{Construct the pseudo-rehearsal buffer.}
Now that we have a model that can generate question-answer pairs for each seen task $t$ and arbitrary images, we can use the learned projectors and the images of the current task dataset $D_t$ to generate questions. Specifically, we can generate a set of pseudo question-answer pairs $\bar{D}_i$ for each task $i<t$ as:
\begin{equation}
    \label{eq:gen-fake-t}
    \bar{D}_i = \{(x_t, \text{split}_?(f^i_{qa} (x_t, p))) \mid (x_t,q_t,a_t) \in D_t\},
\end{equation}
{
where $\text{split}_?(\cdot)$ means we divide the generated text containing the concatenated question and answer based on the position of the symbol 
"?".} Note that $q_t$, $a_t$ are not used when producing the samples.

From the set of generated datasets $\range{\bar{D}_1}{\bar{D}_{t-1}}$ we can obtain a buffer $\bar{\mathcal{M}}_t$ by performing balanced sampling across each task. For instance, if the size of $\bar{\mathcal{M}}_t$ is $M$, we can sample $M/(t-1)$ random triplets from each pseudo dataset. However, this strategy does not take into account the inherent biases of the dataset on the type of questions and the fact that generative models not only reflect but also exaggerate such biases~\citep{naik2023social, zhao2017men}. In the following, we describe how we can populate $\bar{\mathcal{M}}_t$ by taking into account the ideal distribution of question types across each dataset.
\subsection{Pseudo-rehearsal data balancing}
\label{sec:generation-collapse}
Generated data might not reflect the original question-answer distribution and concentrate on most present type of questions/answers, hindering the model generalization capabilities. 
\begin{figure}[htbp] 
    \centering 
    \includegraphics[width=0.95\linewidth]{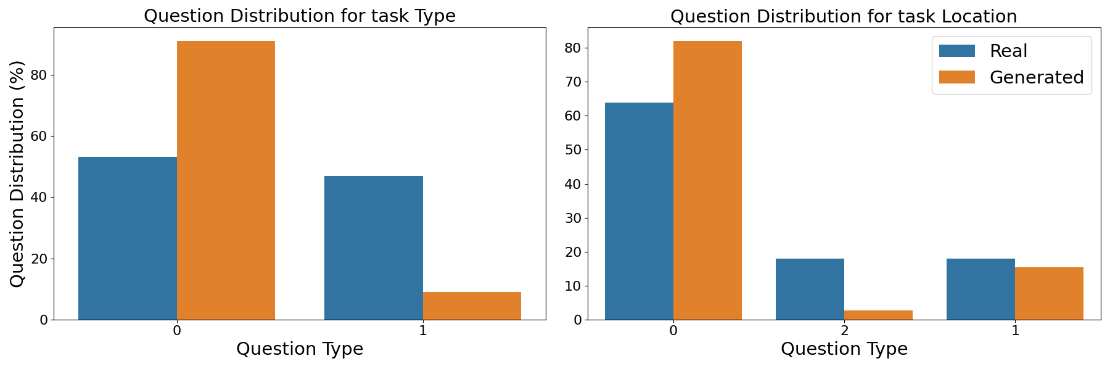}
    \caption{Data-bias present in real vs generated data. Generated questions are heavily skewed towards a certain type of questions. \textbf{Left:} question distribution for task \textit{type}. \textbf{Right:} question distribution for task \textit{location}.}
    \label{fig:ques_dist}
\end{figure}
A significant challenge arises due the tendency of the question generation module to overfit when the training data is unbalanced towards certain question types. This can result in generated question-answer pairs that are skewed towards specific categories, limiting the model’s ability to generalize across the diverse array of questions in the datasets.
As can be seen from Fig.~\ref{fig:ques_dist}-left for the VQACL-VQAv2 benchmark on the \textit{type} task, we observe significant disparities in the distribution of question types. On real data, the questions beginning with "what kind" (type 0) constitute 55.1\% and those starting with "what type" (type 1) make up 44.9\%. However, in the generated data, the distribution is heavily skewed, with "what kind" questions soaring to 90.94\%, and "what type" questions drastically reduced to 9.06\%. Similarly, the generated questions for the task \textit{location} (Fig.~\ref{fig:ques_dist}-right) under-represent questions of type 2, concentrating to the most frequent type 1 ones. This skew illustrates the potential for substantial shift in question generation, underscoring the need for mechanisms to ensure a more uniform distribution of question types to enhance model robustness.
\paragraph{Mitigating the skewed distribution in generation.}
We tackle the generated data shift by employing a post-generation alignment with respect to ground truth data. Let $\mathcal{Q}$ be the space of questions. In an inital step, we define a partitioning function $c_t: \mathcal{Q} \rightarrow \range{1}{K_t}$ mapping questions $q_t^i \in \mathcal{Q}$ associated to task $t$ 
into the associated question type $k, k=\range{1}{K_t}$. Hence we use the same partitioning function $c_t(\cdot)$ to categorize ground truth data $D_t$ and the generated questions $\bar{D}_{t}$. We denote as ${D}_t^k$ the subset of questions in  ${D}_{t}$ associated to type $k$, i.e., ${D}^k_{t} = \{c_t(q) = k | (x,q,a)\in D_t\}$ and with $\bar{D}_t^k$ the generated counterpart. 

From the partitioning, we obtain the expected (ground truth) distribution of question type $k$ as the cardinality of ${D}^k_{t}$ w.r.t. the other types, i.e., $p_t^k = |{D}^k_{t}|/\sum_{j=1}^{K_t}|{D}^j_{t}| $. We can now populate the memory buffer following the obtained distribution. Specifically, let us assume we are at task $t$ and we aim to allocate $\hat{m} = \hat{M}/(t-1)$ number of samples per task, with $\hat{M} < M$. Given the past task $s$ and the set question type $k, k=\range{1}{K_{s}}$, we take a subset $\hat{\mathcal{M}}_s^k$ of random samples from $\bar{D}_s^k$ whose cardinality is $|\hat{\mathcal{M}}_{s}^k| = \ceil{p_{s}^k \cdot \hat{m}}$. The final buffer is the union of all old task and types specific buffers, i.e., $\hat{\mathcal{M}}_{t} = \bigcup_{s=1}^{t-1}\bigcup_{k=1}^{K_s} \hat{\mathcal{M}}_s^k$. In this way, $\hat{\mathcal{M}}_{t}$ follows the expected question type distributions, and can be employed as substitute memory buffer without the need for (i) storing past data (just their meta-statistics) and (ii) biasing the model towards specific question types.
\paragraph{Partitioning function.}
A key component of our approach is the partitioning function $c_j(\cdot)$ used to split real and generated questions into different types. In this paper, we implement two different variants of the partitioning function, denoted as \textit{\OURMODELCLA{}} and \textit{\OURMODELCLU{}}. \OURMODELCLA{} leverages auxiliary meta-information on the question types that is usually available for continual VQA benchmarks. In an initial step, a classifier is trained on task data to predict question types, later on its predictions allow to partition generated data in a coherent manner.
As such meta-information can be unavailable, we also presents \OURMODELCLU{} which clusters data in an unsupervised manner and uses the obtained centroids to apply a coherent partitioning of the generated data buffer. We evaluate both variants in Sec.~\ref{sec:exp:comparison}, as well their performance w.r.t. the buffer size in Sec.~\ref{sec:exp:ablation}.  


\subsection{Sequential learning with pseudo-rehearsal}
\label{sec:sequential-learning}
At each task $t$, the VLM jointly learns to replicate question-answer pairs of the current task while training the VQA head to correctly answer asked questions. 
The data-free pseudo rehearsal buffer avoids forgetting of previous tasks. 
The objective with the pseudo-rehearsal for continual VQA is:
\begin{equation}
    \label{eq:rehearsal-generic}\mathcal{L}^t_{\text{CL}} = \frac{1}{N_t+\hat{M}}\sum_{(x,q,a)\in R_t} \mathcal{L}_{\text{lm}}(f(x,q), a),
\end{equation}
where $R_t=D_t\cup \mathcal{\hat{M}}_t$ with 
$\mathcal{\hat{M}}_1 = \emptyset$, and $f$ denotes the learning VQA model using the answer language projector $f_{v\to a}$. 
 By accounting for the loss regarding question-answer generation $\mathcal{L}_{QA}^t, t=\range{1}{T}$ (defined in Eq.~\ref{eq:qa-objective}), the final loss function at task $t$ can be expressed as:
\begin{equation}
\mathcal{L}^t = \mathcal{L}^t_{CL} + \mathcal{L}_{QA}^t.
\end{equation}
\section{Experiments}
We evaluate \OURMODEL{} on two benchmark datasets that feature VQACL, in comparison to state-of-the-art methods. In the following, we first introduce the datasets, the evaluation protocol and implementation details. Then, we discuss the main comparison conducted on the two datasets. Finally, we provide the ablation study to justify our method design.

\paragraph{Datasets.} We consider two benchmark datasets, namely VQACL-VQAv2~\citep{zhang2023vqacl} and CLOVE-function~\citep{lei2023symbolic}.
VQACL-VQAv2 builds on VQA v2~\citep{goyal2017making} and it constructs a dual-level task sequence where target objects (\ie, the subject of the questions) and performed tasks are nested together. 
At the outer level, $T=10$ tasks are defined based on different skills, \eg, location recognition, general recognition, counting, commonsense, etc. Within each task, skills are applied to a sequence of 5 different object groups, leading to an inner continual learning setting based on target objects. 
We focus on the standard testing evaluation, where test data comes from the same distribution as training data. 
CLOVE-function benchmark gathers data from GQA~\citep{hudson2019gqa}, CRIC~\citep{gao2022cric} and TextVQA~\citep{singh2019towards}, defining a sequence of 6 tasks about different properties: objects, attributes, relations, logical, knowledge, and scene-text. Note that for the scene-text task, the method proposed in~\citep{lei2023symbolic} leverages specialized backbone and features for the OCR queries. Thus, to avoid unfair comparison, we leave out the scene-text task, and hence, consider a setting with the remaining $T=5$ tasks. We refer to the Supp. Mat. for more details.  

\paragraph{Performance metrics.} Following prior works~\citep{zhang2023vqacl, smith2023construct}, we evaluate the approach in terms of final model performance and forgetting of previous tasks. Specifically, we measure the average performance (AP) at the end of the CL sequence and the average forgetting (AF)~\citep{chaudhry2018efficient, lopez2017gradient} that probes the average model degradation due to learning of subsequent tasks. 
Formally, let $a_{i,j}$ be the performance of the model trained on task $i$ and evaluated on task $j$. We define the average performance as $AP=\frac{1}{T}\sum_{t}a_{T,t}$ ($\uparrow$), while the average forgetting is evaluated as $AF= \frac{1}{T-1}\sum_t max_{t\leq i\leq T-1} a_{i, t} -  a_{T, t}$ ($\downarrow$).

\paragraph{Implementation details.} 
We implement our method with BLIP-2~\citep{li2023blip}, a VLM pre-trained to bridge the vision-language modality gap. Note that our approach can generalize to other VLMs with similar functional components, such as LLaVa~\citep{liu2024visual} or MiniGPT-4~\citep{zhu2023minigpt}. 
During adaptation, we freeze the model and limit fine-tuning to language projection layers only. 
We use Adam optimizer with a fixed learning rate of \(1 \times 10^{-4}\) where the best validation checkpoint is resumed for continual training. The training follows the same hyperparameter settings as respective benchmarks.
We consider an initial generated buffer of size $M=20\text{k}$, later downsampled to $\hat{M}=5\text{k}$, where samples are balanced according to features extracted from a frozen \texttt{all-mpnet-base-v2}. While \OURMODELCLA{} classifies questions according to the available meta-information, for \OURMODELCLU{} we select the number of clusters to be 7 and 10 for VQACL-VQAv2 and CLOVE-function respectively, according to validation performance. See Supp. for more implementation details.



\subsection{Comparison}\label{sec:exp:comparison}
\paragraph{Baselines.} 
For fair comparison, we implement all baselines with the BLIP-2~\citep{li2023blip} backbone as \OURMODEL{}. We consider the off-the-shelf application of the pre-trained model (\texttt{Zero-shot}) and basic learning strategies, such as naive sequential learning (\texttt{Seq-FT}) and the joint training that considers all data available at once (\texttt{Multi-task}). 
In addition, we compare with continual learning regularization strategies, including EWC~\citep{kirkpatrick2017overcoming} and MAS~\citep{aljundi2018memory}, a rehearsal strategy replaying previous tasks data (\texttt{Rehearsal}) and state-of-the-art prompting strategy L2P~\citep{wang2022learning}. We further adapted the text-based LAMOL (denoted with * for the adaptation) to the data-free VQACL setting, where a single language head leverages current task images to jointly train for question-answer generation and VQA. Finally, we compare \OURMODEL{} with SOTA continual VQA strategies such as VQACL~\citep{zhang2023vqacl} on the VQACL-VQAv2, and SGP~\citep{lei2023symbolic} on CLOVE-function.
\begin{table}[t!]
\caption{The continual learning performance in terms of AP and AF on the VQACL-VQAv2 dataset. All methods use BLIP-2 as their backbone. Best data-free method is highlighted in bold.}
\centering
\resizebox{0.35\textwidth}{!}{%
\begin{tabular}{lccc}
\toprule
\textbf{Method} & \textbf{Data-free} &\textbf{AP} ($\uparrow$) & \textbf{AF} ($\downarrow$)\\
\midrule
Zero-shot  & - & 40.84 & - \\
\hdashline
Rehearsal & \xmark & 46.94 & 4.21 \\
Multi-task & \xmark &49.55 & -\\
VQACL~\citep{zhang2023vqacl} & \xmark & 49.80 & 1.18 \\
\hdashline
Seq-FT & \cmark &41.29 & 15.98 \\
EWC~\citep{kirkpatrick2017overcoming} & \cmark & 40.62 & 9.78\\
MAS~\citep{aljundi2018memory} & \cmark &40.27 & 10.98 \\
L2P~\citep{wang2022learning} & \cmark & 40.15 & 16.56 \\
LAMOL*~\cite{sun2019lamol} & \cmark & 28.26 & 6.81 \\
\textbf{\OURMODELCLA~(Ours)} & \cmark & 47.65 & 3.61 \\
\textbf{\OURMODELCLU~(Ours)} & \cmark & \textbf{48.40} & \textbf{1.40} \\
\bottomrule
\end{tabular}%
}
\label{tab:vqacl-short}
\end{table}

\paragraph{Discussion.} Table \ref{tab:vqacl-short} presents the results on the \textit{VQACL-VQAv2} benchmark dataset. 
Notably, the Zero-shot BLIP-2 shows competitive with continual learning baselines, demonstrating the inherent effectiveness of the BLIP-2 backbone for the task. 
Our proposed~\OURMODELCLU{}, which incorporates balancing the rehearsal buffer using clustering,  achieves an AP of 48.40, slightly outperforming our classifier-based variant ~\OURMODELCLA{} with 47.65 AP.
Intuitively, the classifier controls the alignment of the generated data based on the single-labelled property of the question. On the contrary, clustering-based grouping allows for accounting for different relevant aspects of the questions. Surprisingly, these results surpass the rehearsal strategy involving real data samples. The recent method VQACL and the Multitask baseline are only marginally better than \OURMODEL{}, despite having access to past tasks real data. 
Our methods effectively reduce knowledge loss, as demonstrated by the lower AF rate of 1.40 for \OURMODELCLU{} compared to the 4.21 AF rate of the Rehearsal method. Under the same memory constraint, LAMOL* achieves a low 28.26\% AP, indicating its unsuitability for the setting. Further inspection on generated questions shows that the single language projection layer fails to effectively rehearse question generation for old tasks.
%
In line with recent findings ~\citep{smith2023construct}, continual learning strategies are ineffective in managing the trade-off between adaptation to the different tasks and preservation of original knowledge, as showed by the lower performance with respect to the zero-shot evaluation. 
On one hand, L2P mitigates the catastrophic forgetting on the vision side only, without accounting for the multimodal nature of the problem. On the other, regularization approaches fail to estimate the relevance of weights. We hypothesize the text generation head does not provide a suitable space to estimate how knowledge is stored. Even the best performing regularization method, EWC, only achieves 40.62 in terms of AP.
\begin{table}[t!]
\caption{The continual learning performance in terms of AP and AF on the CLOVE-function dataset. Best data-free method in bold.}
\centering
\resizebox{0.46\textwidth}{!}{%
\begin{tabular}{lcccc}
\toprule
\textbf{Method} & \textbf{Backbone} &\textbf{Data-free} & \textbf{AP} ($\uparrow$) & \textbf{AF} ($\downarrow$)\\
\midrule
Zero-shot & BLIP-2 & - & 24.47 & - \\
\hdashline
Rehearsal & BLIP-2 & \xmark & \textbf{41.82} & 3.14 \\
Multi-task  & BLIP-2 & \xmark & 32.26 & - \\
\hdashline
SGP~\cite{lei2023symbolic}  & UniVQA & \cmark & 27.72 & 38.16 \\
Seq-FT  &BLIP-2 & \cmark & 22.70 & 22.19\\
EWC~\citep{kirkpatrick2017overcoming} &BLIP-2 & \cmark & 23.90 & 10.14 \\
MAS~\citep{aljundi2018memory}  & BLIP-2 & \cmark & 26.30 & 6.46 \\
L2P~\citep{wang2022learning}  &BLIP-2 & \cmark & 16.86 & 16.02 \\
LAMOL*~\cite{sun2019lamol} & BLIP-2 & \cmark & 15.26 & 10.46\\
\textbf{\OURMODELCLA{} (Ours)}  & BLIP-2 & \cmark & 38.34 & \textbf{2.85} \\
\textbf{\OURMODELCLU{} (Ours)} & BLIP-2 & \cmark & 36.57 & 4.61 \\
\bottomrule
\end{tabular}%
}
\label{tab:clove-short}
\end{table}

Table~\ref{tab:clove-short} reports the results on \textit{CLOVE-function} benchmark. CLOVE-function raises different challenges with respect to VQACL-VQAv2, such as the more diverse set of tasks, the compositional nature of questions, and the needed auxiliary information required for the knowledge task. Overall, we observe similar trends as in Table~\ref{tab:vqacl-short}.
Large pre-trained models benefit the final task performance in the continual learning setting, as evidenced by the strong performance of the zero-shot BLIP-2 model, compared with the state-of-the-art specialized architecture UniVQA.
\OURMODEL{} achieves 38.34 in terms of AP, with the cluster variant being closely behind at 36.57 AP. Both variants of our \OURMODEL{} approach achieve the best performance among data-free methods, only lagging behind the real rehearsal method by a mere $3\%$.
Our model significantly bridges the gap between rehearsal and data-free approaches, requiring a minimal reliance on real data from past tasks.
Moreover, our model demonstrates a strong capability in handling diverse and complex question types, typical of CLOVE-function, as demonstrated by the 32.26 AP of the Multitask baseline. We conjecture the negative transfer across tasks hinders the model generalization capabilities. \OURMODEL{} is effective especially in challenging relational and logic-based queries (see the Supp. Mat. for extended results on all tasks).
Interestingly, while EWC achieves a performance similar to the Seq-FT baseline at 22.70 AP, MAS yields a 26.60 AP proving the most effective approach among the classic continual learning strategies. 
L2P instead confirms the unsuitability of the approach to the multimodal setting~\citep{smith2023construct}.
\subsection{Ablation studies} \label{sec:exp:ablation}
We study the different components and design choices of \OURMODEL{}, by analyzing the impact of various balancing strategies, question category conditioning, pseudo-labelling and the images used to populate the memory. We perform the ablation with VQACL-VQAv2 and analyze the results below.

\begin{figure}[t!]
    \centering
    \includegraphics[width=0.66\linewidth]{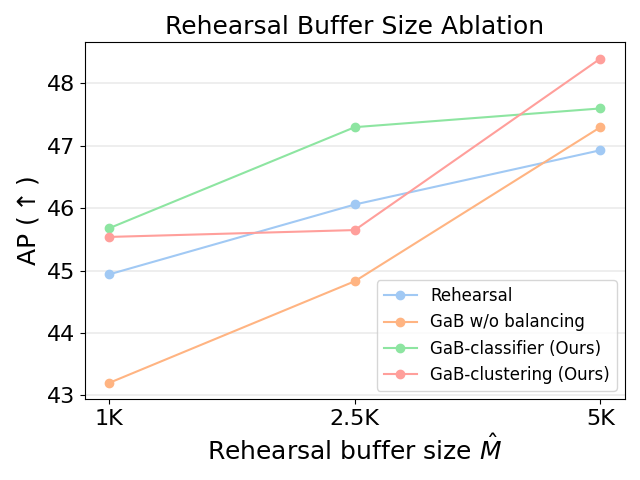}
    \caption{Analysis of balancing strategies at varying rehearsal buffer sizes (1k, 2.5k, and 5k samples) on the VQACL-VQAv2 benchmark in terms of AP.}

    \label{fig:balancing_ablation}
\end{figure}

\begin{figure}[t!]
    \centering
\includegraphics[width=1.\linewidth]{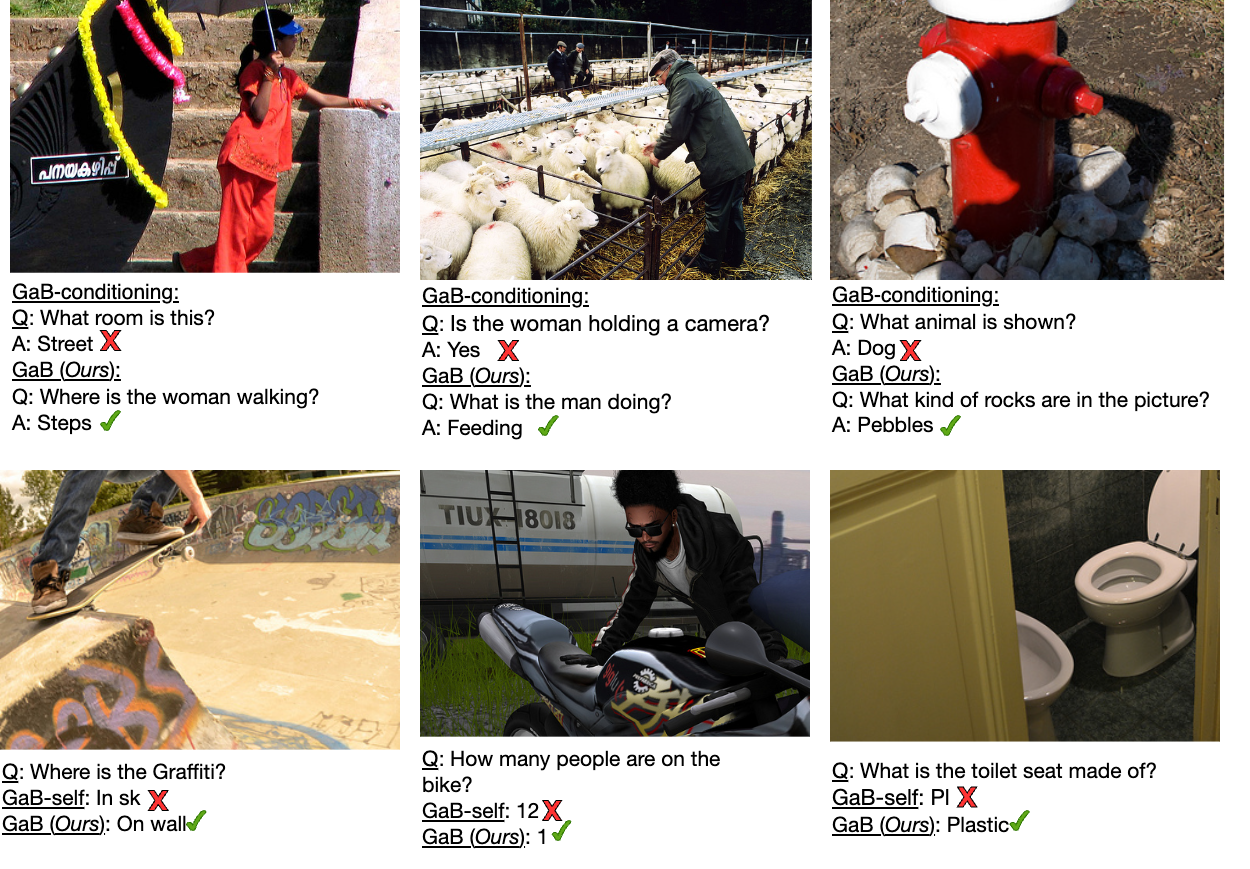}
   \caption{Qualitative visualization of the generated pseudo-rehearsal data on the VQACL-VQAv2 benchmark. \textbf{Top row:} analysis on generation conditioning for questions balancing. \textbf{Bottom row:} analysis on the pseudo-strategy used for answer generation.}
    \label{fig:vqacl-qualitative-pseudo-questions}
\end{figure}

\paragraph{Balancing strategies and the buffer size.}
To assess the impact of our balancing strategy, we experiment with a variant, referred as \texttt{\OURMODEL{} w/o balancing}, where we remove the balancing component. We also analyze the impact of varying sizes of the rehearsal buffer on the variants of our methods. 
As depicted in Figure \ref{fig:balancing_ablation}, both the \OURMODELCLA{} and \OURMODELCLU{} methods improve performance with increasing buffer sizes. The results show that our balancing strategies significantly outperform the scenarios where no balancing is applied, the gain is consistent across all considered buffer sizes. Notably, with the small-sized buffers, \OURMODELCLA{} and \OURMODELCLU{} yields a similar performance, providing a benefit with respect to the rehearsal strategy with real data. As the buffer size increases to $\hat{M}=5\text{k}$ samples, \OURMODELCLU{} emerges as the most effective, achieving the highest AP, highlighting the scalability  of the clustering approach under conditions of larger data availability.
\paragraph{Question category conditioning.}
\OURMODEL{} generates questions freely, based on the image content. As we discussed in Sec.~\ref{sec:generation-collapse}, this results 
in a skewed distribution towards more prevalent question types and we need to fix this issue via a 
post-generation balancing strategy. A different direction is to avoid this problem altogether by  
conditioning the question generation process 
on target question types. 
We study this alternative strategy, 
and compare the proposed approach with \texttt{GaB-conditioning} where the language model is trained by explicitly prompting to generate questions based on their type, p="\texttt{Question type <type>: Question: <question> Answer: <answer>}".
During inference, question generation is similarly driven by the classifier partitioning function to ensure a balanced distribution. 
Table \ref{tab:clove-ablation} presents the results of this conditioning approach in comparison to our methods involving only post-generation processing. 
We observe that the results of  \OURMODEL-conditioning are inferior to ours, going from 47.65 to 44.07 in terms of AP. 
Upon further inspection, we find that forcing the model to produce questions related to certain types can lead to false assumptions, such as forcing to ask a question type that is not relevant for the given image (see Fig. \ref{fig:vqacl-qualitative-pseudo-questions} for illustration). 

\paragraph{Pseudo-labelling strategy.}
In our method, rehearsal generation simultaneously produces question-answer pairs based on image inputs. Alternatively, we explored using our trained answer generation model to respond to these questions separately in a self-training manner (\texttt{\OURMODEL-self}). As detailed in Table \ref{tab:clove-ablation}, this method resulted in a significant decrease in AP by approximately 2.65, likely due to the inferior quality of the pseudo-answers, which are often noisy or incomplete. These issues may cause confirmation bias during fine-tuning. For illustrative examples, see the bottom row of Fig.~\ref{fig:vqacl-qualitative-pseudo-questions}.
.
\begin{table}[t!]
\caption{Analysis on the design choices of \OURMODEL{} in terms of AP and AF on the VQACL-VQAv2 dataset.}
\centering
\resizebox{0.25\textwidth}{!}{%
\begin{tabular}{lcc}
\toprule

\textbf{Method} & \textbf{AP}($\uparrow$) & \textbf{AF} ($\downarrow$)\\
\midrule
\textit{Generation conditioning} & & \\
\OURMODEL-conditioning  & 44.07 & 6.45 \\
\addlinespace
\textit{Pseudo strategy} & & \\
\OURMODEL-self & 45.01 & 4.76 \\
\addlinespace
\textit{Rehearsal on Task data} & & \\
\OURMODEL-pastimages & 38.78 & 4.07 \\
\midrule
\textbf{\OURMODELCLA{} (Ours)} & \textbf{47.65} & \textbf{3.61} \\
\addlinespace
\bottomrule
\end{tabular}%
}
\label{tab:clove-ablation}
\end{table}
\paragraph{Using past visual images.}
In this study, we evaluate the effectiveness of generating question-answer (QA) pairs with images from the current versus previous tasks. Our goal is to discern why our rehearsal strategy, which uses current task images for QA generation, surpasses traditional methods utilizing real-data. In Table \ref{tab:clove-ablation}, we demonstrate with \texttt{\OURMODEL-pastimages} that applying our QA generation strategy to past task images results in a performance decrease of about 10\% compared to using current task images, as shown in \OURMODELCLA{}.

We believe that using current task images for QA pair generation offers enhanced contextual relevance, similar to data augmentation. This approach not only provides more context by posing multiple questions on the same images but also helps the model bridge knowledge gaps across tasks, effectively serving as a form of positive regularization. Consequently, the model better assimilates new information with previously acquired knowledge, improving its ability to maintain relevant capabilities across different tasks.

\section{Conclusion}
We presented \OURMODEL{}, a novel data-free method for VQACL building on top of the inherent generative capability of VLMs. 
With current task images, \OURMODEL{} synthesizes data samples of previous VQA tasks by learning to generate visually conditioned question-answer pairs. We tackled the model generation issue where the questions are skewed to the most-occurring questions, by balancing the generated data according to real data statistics or patterns. Results on standard benchmarks, including VQACL-VQAv2 and CLOVE-function, prove the benefit of~\OURMODEL{} against existing data-free methods. 
As \OURMODEL{} depends on the data generated for rehearsal, noisy or ambiguous pseudo-samples can hinder the model performance. Future work will focus on improving generated data quality, introducing techniques such as hallucination mitigation and automated quality control.

\section{Acknowledgment}
The authors acknowledge the CINECA award under the ISCRA initiative for the availability of high-performance computing resources and support. Deepayan Das is supported by the PRIN project “B-FAIR” (Prot.2022EXF3HX) and the PAT project “AI@TN". This work was
supported by the projects EU Horizon ELIAS (No. 101120237), AI4TRUST (No.101070190), FAIR Future AI Research (PE00000013), funded by NextGeneration EU, and carried out in the Vision and
Learning joint laboratory of Fondazione Bruno Kessler and the University of Trento, Italy

\clearpage
{\small
\bibliographystyle{ieee_fullname}
\bibliography{references}
}

\end{document}


\title{Supplementary Material: \\One VLM to Keep it Learning: Generation and Balancing for Data-free Continual Visual Question Answering}

\author{Deepayan Das\textsuperscript{$1$} \quad
Davide Talon\textsuperscript{$2$} \quad
Massimiliano Mancini\textsuperscript{$1$} \quad
\\
Yiming Wang\textsuperscript{$2$} \quad
Elisa Ricci\textsuperscript{$1,2$} \\
\small
$^1$University of Trento \quad \quad $^2$Fondazione Bruno Kessler \\
\small
\texttt{\{deepayan.das, massimiliano.mancini, e.ricci\}@unitn.it}\\
\small
\texttt{\{dtalon, ywang\}@fbk.eu}
}
\maketitle

\appendix
\section{Supplementary Overview}
In this supplementary material we present further results and ablation analysis on the presented method \OURMODEL{},  implementation details and qualitative visualization of generated question-answer pairs. Specifically, while in Section~\ref{sup:extended-results}, we report extensive results on the method, Section~\ref{sup:ablation} ablates the proposed pseudo-rehearsal strategy. We continue with Section~\ref{sup:dataset} and Section~\ref{sup:implementation} describing in further detail the datasets and the implementation details of the approach, respectively. Finally, Section~\ref{sup:qualitatives} concludes by providing qualitatives of generated question answer pairs.
\section{Extended Results}
\label{sup:extended-results}
We here report extended empirical results for the proposed method \OURMODEL{}, including the intermediate evaluation of the sequentially training model and the robustness of the approach to different task orders.
\subsection{Per-Task Performance Analysis}

We evaluate the performance of the learning VQA model on the different tasks as training progresses.

\begin{figure*}[htbp]
\centering
\includegraphics[width=\linewidth]{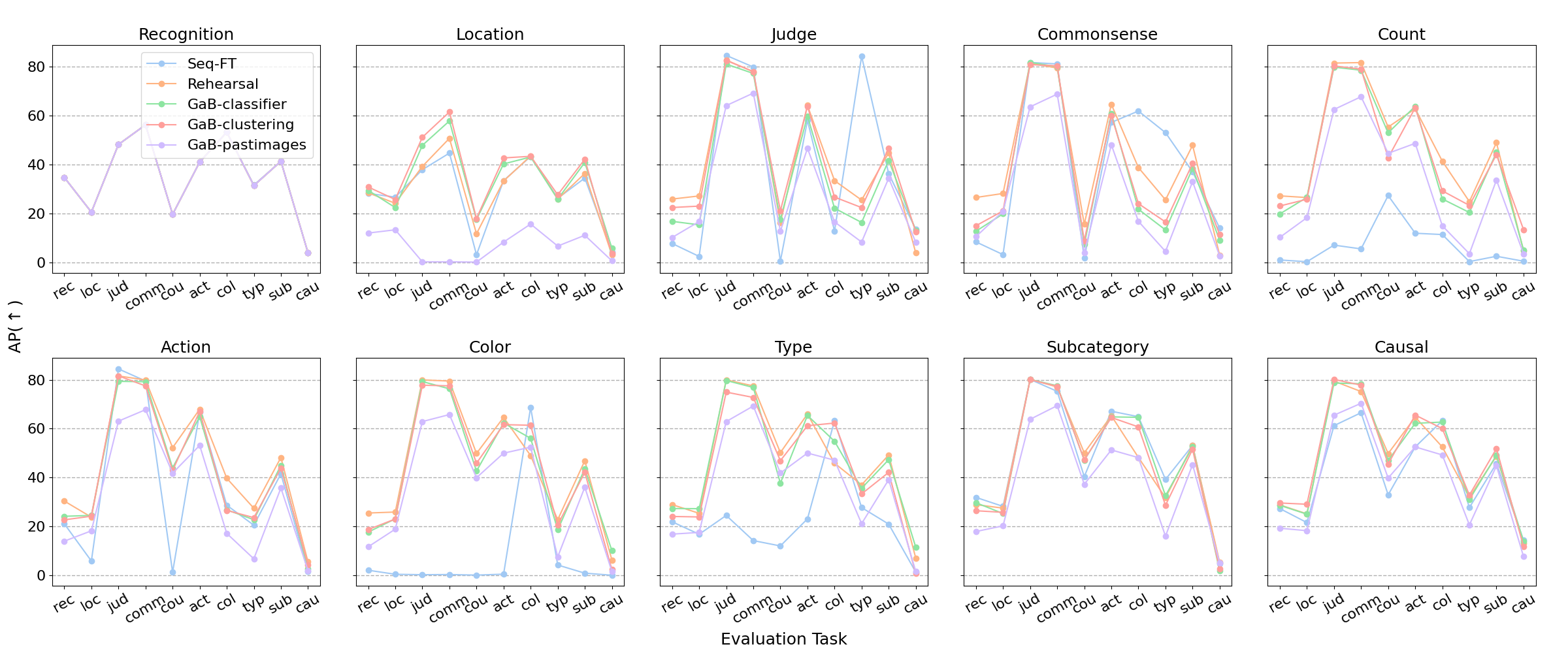}
\caption{Per-task performance in terms of AP across different tasks in the VQACL-VQAv2 benchmark.}
\label{fig:vqacl_performance}
\end{figure*}

\begin{figure*}[htbp]
\centering
\includegraphics[width=\linewidth]{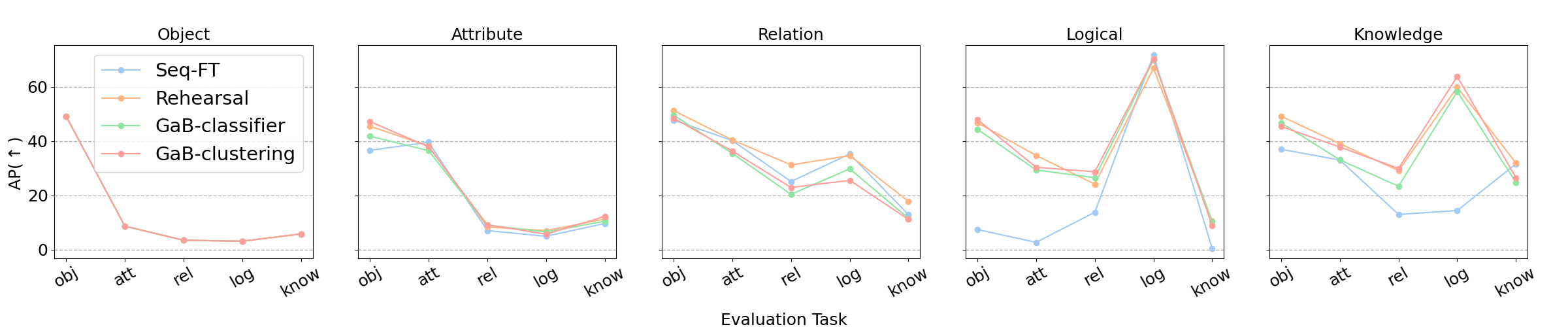}
\caption{Per-task performance in terms of AP across different tasks in the CLOVE-function benchmark.}
\label{fig:clove_performance}
\end{figure*}
Figure \ref{fig:vqacl_performance} showcases the performance variations across various tasks within the VQACL-VQAv2 benchmark in terms of AP, highlighting how each continual learning method adapts over time. As can be noted, despite building on pseudo-rehearsal samples only, throughout the entire sequential adaptation \OURMODEL{} achieves accuracy on par with the rehearsal strategy that leverage past real data. A similar behaviour is observed in Figure \ref{fig:clove_performance} detailing the sequential task performance for the CLOVE-function benchmark during sequential training.
\subsection{Continual Learning Across Task Orders on CLOVE}

Continual learning performance can significantly vary depending on the order in which tasks are presented. To explore this variability, we evaluate \OURMODEL{} across three different task orders of the CLOVE-function benchmark. Let us  denote each task with its initial letter, \ie, Objects (o), Attributes (a), Relations (r), Logical (l) and Knowledge (k), we consider task orders `oarlk', `rolak', and `lkora'.

\begin{table}[htbp]
\caption{The continual learning performance in terms of AP and AF on the CLOVE-function dataset considering different task orders.}
\centering
\resizebox{0.45\textwidth}{!}{%
\label{tab:performance_comparison}
\begin{tabular}{lcccccc}
\toprule
 & \multicolumn{2}{c}{oarlk} & \multicolumn{2}{c}{lkora} & \multicolumn{2}{c}{rolak} \\
\cmidrule(lr){2-3} \cmidrule(lr){4-5} \cmidrule(lr){6-7}
Method & AP ($\uparrow$) & AF ($\downarrow$) & AP ($\uparrow$) & AF ($\downarrow$) & AP ($\uparrow$) & AF ($\downarrow$)\\
\midrule
Multi-task & \multicolumn{6}{c}{32.26}\\
\hdashline
Rehearsal       & 41.82 & 3.14 & 29.75 & 13.61 & 28.23 & 8.57 \\
Seq-FT          & 22.70 & 22.19 & 13.35 & 24.24 & 12.35 & 23.61\\
GaB w/o balancing & 37.01 & 3.61 & 25.48 & 16.9 & \textbf{29.18} & \textbf{7.97} \\
\textbf{\OURMODELCLU{} (Ours)} & \textbf{40.70} & \textbf{1.40} & \textbf{31.59} & \textbf{10.64} & 26.17 & 9.67\\
\bottomrule
\end{tabular}}
\label{table:task order}
\end{table}

The results are summarized in Table \ref{tab:performance_comparison} in terms of Average Performance (AP) and Average Forgetting (AF). Despite the differences in AP and AF across the task sequences, similar behaviours are observed: our approach \OURMODELCLU{} performs competitively with the rehearsal strategy and it provides a large margin improvement to the Seq-FT baseline. Notably, the `oarlk' sequence consistently shows better performance metrics compared to `lkora' and `rolak', suggesting that the order in which tasks are encountered can influence the efficacy of the learning process. In the `oarlk' sequence, our \OURMODELCLU{} method demonstrated the best resilience against forgetting with an AF of 1.40, and a high AP of 40.70, underscoring its robustness in handling the challenges presented by this particular sequence. We see a similar trend in the sequence `lkora' where our method has the best overall AP and AF. Conversely, in the `rolak', GaB w/o balancing shows the most effective, indicating that pseudo-generation aids forgetting mitigation while the clustering-based balancing strategy might require more careful hyperparameter tuning, such as the number of clusters to use.

\subsection{Balancing Questions}
 
We visualize the question types distribution of rehearsal samples before (\texttt{Generated}) and after balancing (\texttt{Balanced}) with our pseudo-rehearsal balancing module, compared to the ground truth one (\texttt{Real}). These visualizations help illustrate the impact of our balancing technique on the diversity of question types generated during the pseudo-rehearsal data generation.

\begin{figure*}[h]
\centering
\includegraphics[width=1\linewidth]{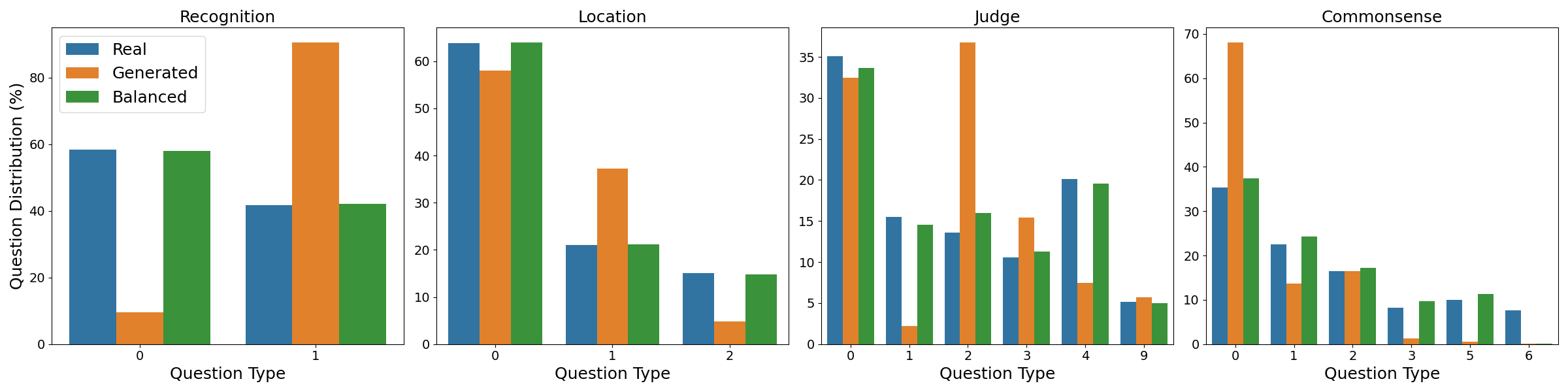}
\caption{Question distribution before and after pseudo-rehearsal balancing for the VQACL-VQAv2 benchmark. The figure shows the distribution across different question categories for old tasks (as per plot title) generated from last task \emph{count} visual images.}
\label{fig:ques_dist_vqacl}
\end{figure*}

\begin{figure*}[t]
\centering
\includegraphics[width=\linewidth]{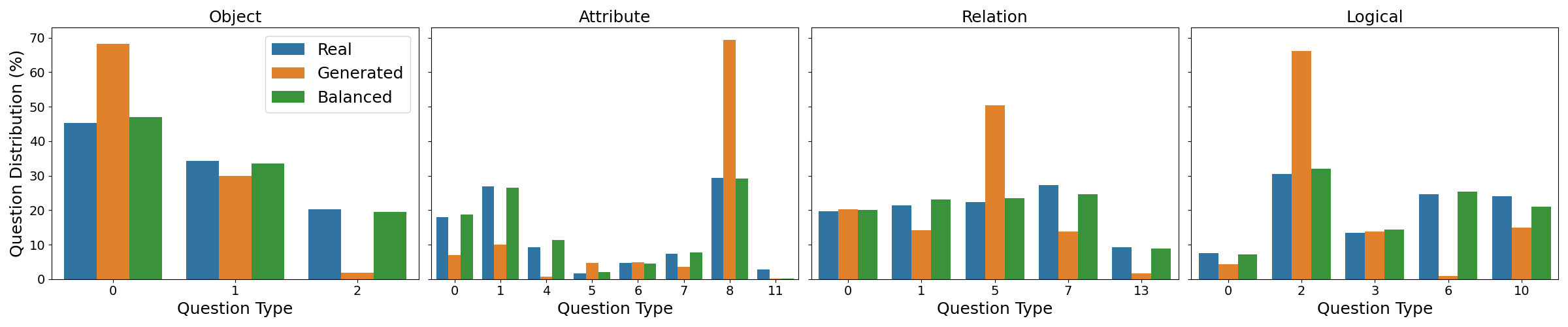}
\caption{Question distribution before and after pseudo-rehearsal balancing for the CLOVE-function benchmark. The figure shows the distribution across different question categories for old tasks (as per plot title) generated from last task \emph{knowledge} visual images.}
\label{fig:clove_question_dist}
\end{figure*}
Figure \ref{fig:ques_dist_vqacl} illustrates the distribution alignment for VQACL-VQAv2 demonstrating how \OURMODELCLU{} ensures no single question type dominates the training process. We show similar results on CLOVE-function in Figure~\ref{fig:clove_question_dist}.

\section{Further ablations}
\label{sup:ablation}
We further ablate \OURMODEL{} on the number of balancing clusters and the use of the question-answer generation module for dynamic sampling of the rehearsal data. 

\begin{figure}[h]
    \centering
    \includegraphics[width=0.8\linewidth]{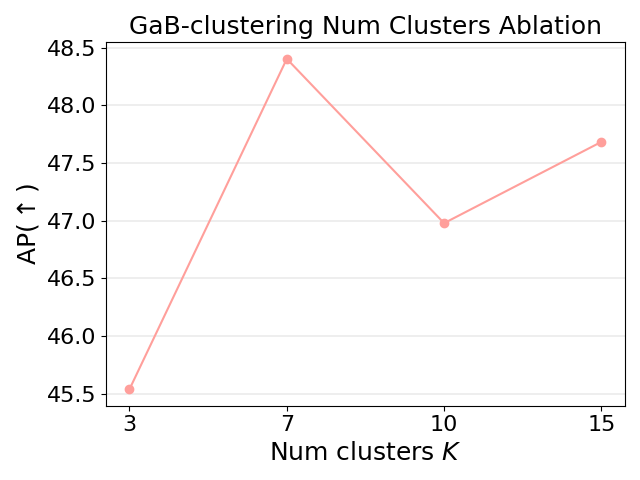}
   \caption{ The continual learning performance of \OURMODELCLU{} in terms of AP when varying the number of balancing clusters $K$ on the VQACL-VQAv2 dataset.}
    \label{fig:n_clusters_ablation}
\end{figure}

\subsection{Varying the number of balancing clusters}
We explore the impact of varying the number of clusters in our balanced cluster strategy~\OURMODELCLU. We ablate to determine the optimal number of clusters that yields the best performance. The results of this study are illustrated in Figure \ref{fig:n_clusters_ablation}, which displays the average precision achieved across different cluster counts $K$. Our findings indicate that setting the number of clusters to 7 maximizes average perfomance.

\subsection{Dynamic Pseudo-Rehearsal}
Traditional rehearsal strategies for continual learning are constrained in the buffer dimension due to the limited availability of question-answers from previous tasks. Thanks to the dynamic generation of samples, pseudo-rehearsal strategies could possibly rely on larger replay buffers instead. We here explore a dynamic version of the presented approach (\texttt{GaB-dynamic}) where pseudo-samples are generated on-the-fly based on current task data. At each training step this method involves dynamically generating a question related to previous tasks for each current batch visual image. As the alignment of the question distributions is non-trivial in the batch low-samples setting, no balancement to the generated questions-answer pairs is applied.

Table~\ref{tab:clove-short} compares the approaches in terms of AP and AF for the CLOVE-function setting. Despite the larger number of replay samples available to \OURMODEL-dynamic, the results indicate that \OURMODELCLU{} achieves the highest performance with a +4.53\% and -3\% in the AP and AF metrics, respectively. Intuitively, we observe limited variability of QA samples that could be generated within each batch with a consequent drop 
on the final performance.
                                                                                          
\begin{table}[t!]
\caption{The continual learning performance in terms of AP and AF on the CLOVE-function dataset.}
\centering
\resizebox{0.38\textwidth}{!}{%
\begin{tabular}{lcc}
\toprule
\textbf{Method} & \textbf{AP} ($\uparrow$) & \textbf{AF} ($\downarrow$)\\
\midrule
GaB-dynamic & 36.17 & 5.26 \\
\textbf{\OURMODELCLA{} (Ours)} & 37.97 & 5.25\\
\textbf{\OURMODELCLU{} (Ours)} & \textbf{40.70} & \textbf{2.26} \\
\bottomrule
\end{tabular}%
}
\label{tab:clove-short}
\end{table}

\section{Dataset Details}
\label{sup:dataset} 

We provide comprehensive details on both datasets utilized in our study. The VQACL-VQAv2 benchmark is comprised of 10 distinct tasks, each with its own set of questions tailored to specific aspects of visual and textual understanding. The different tasks considered are in order: \texttt{Recognition}, \texttt{Location}, \texttt{Judge}, \texttt{Commonsense}, \texttt{Count}, \texttt{Action}, \texttt{Color}, \texttt{Type}, \texttt{Subcategory} and \texttt{Causal}. We refer the reader to the original paper for extensive details on the benchmark. On the other hand, the CLOVE benchmark includes 6 tasks. For fair comparison, we restrict our evaluation to only 5 tasks due to the specialized architecture and auxiliary features needed for answering scene-text questions asking to OCR present text.

Each task within these datasets is associated with questions that are categorized into various types depending on their content and focus. These question types are labeled systematically to facilitate targeted training and analysis. While in VQACL-VQAv2 meta-information on question types models the initial words used in the question construction, e.g. "what type", "is the", "where is", "how many", differently, CLOVE-function auxiliary information models the property being queried and the expected answer, for instance "MaterialChoose", "activityWho" or "relVerify".




\section{Implementation details}
\label{sup:implementation}
We implement our strategy in PyTorch~\citep{paszke2019pytorch} and employ the Hugging Face\footnote{\url{https://huggingface.co/}} implementation of the BLIP-2\citep{li2023blip} architecture, specifically the \texttt{opt-2.7b} version. For textual generation, we follow standard practice and fix the \textit{max\_new\_tokens} for both generated answers (2 tokens) and pseudo QA pairs (20 tokens). The repetition penalty is set to 1.2. In line with prior work, we prompt BLIP-2 for answer generation with the prompt \texttt{p="Question: <question> Answer:"}, while pseudo-QA has an empty prompt \texttt{p=""}, akin to the original BLIP-2 strategy.
\paragraph{Baselines.} We implemented continual learning strategies following an open-source codebase for CL\citep{lomonaco2021avalanche}. The strategies are applied to the same BLIP-2 architecture as \OURMODEL{}. For the regularization-based methods, Elastic Weight Consolidation (EWC) and Memory Aware Synapses (MAS), we set the regularization parameters to 1.0 and compute importance weights on the token classifier generating the output textual sequence. For the Learning to Prompt (L2P) approach the prompting strategy is applied to the visual encoder only. We adopt different configurations depending on the benchmark: for VQACL-VQAv2, we utilize a prompt pool size of 10 with a regularization parameter of 1.0, whereas for CLOVE-function, the prompt pool size is increased to 50 and the regularization parameter is adjusted to 0.5. These settings are carefully chosen to optimize performance across the diverse conditions presented by each benchmark.

\section{Qualitatives Results}
\label{sup:qualitatives}
\begin{figure*}[h]
    \centering
     \includegraphics[width=1\linewidth]{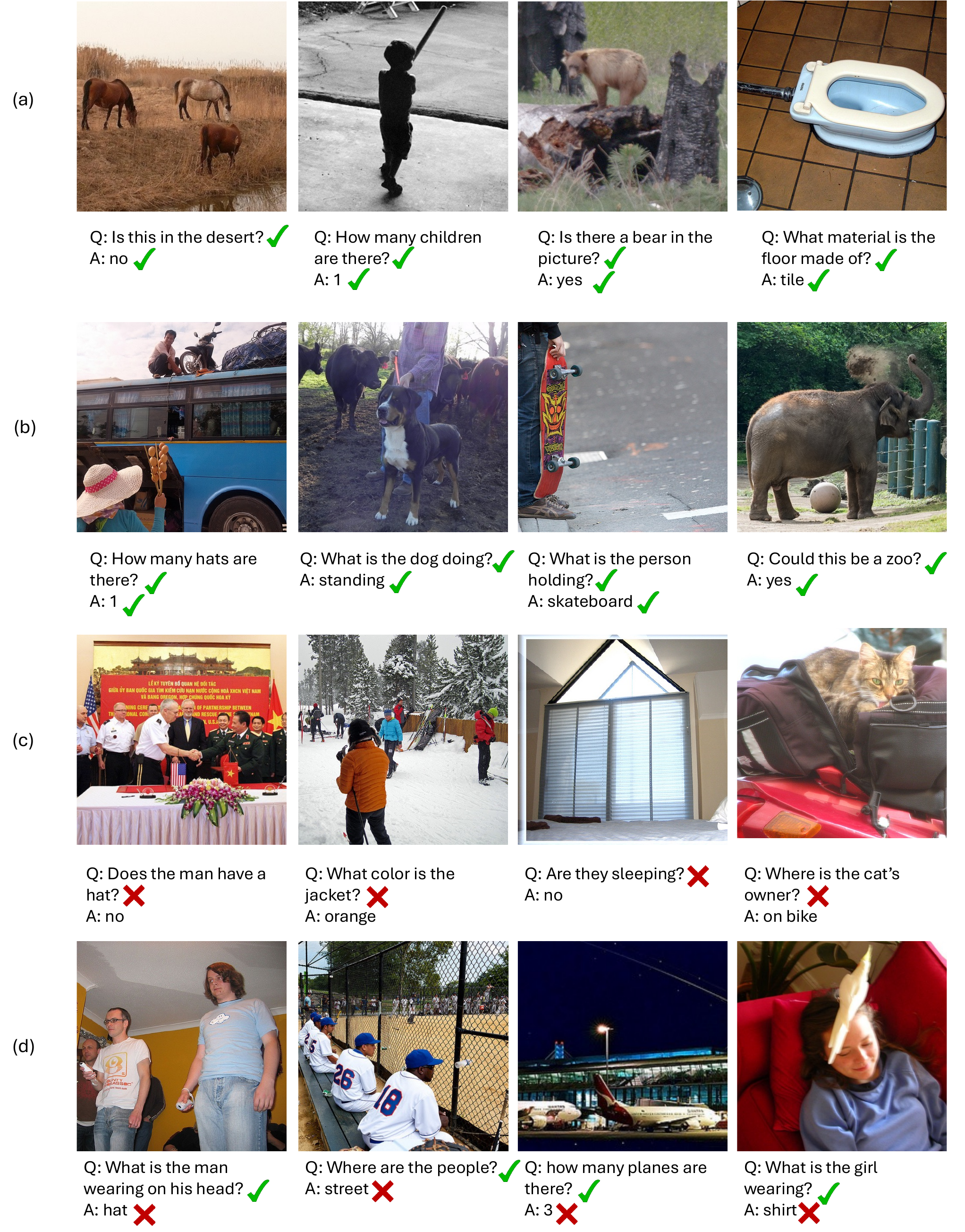}
    \caption{Qualitatative results of GaB generated QA pairs on VQACL-VQAv2 datasets. \textbf{(a)-(b)} Correctly generated questions answer pairs; \textbf{(c)} question-answer pairs with ill-posed questions; \textbf{(d)} question-answer pairs with wrong generated answers.}

\label{fig:supp-vqacl-further-qualitatives}
\end{figure*}

We report qualitative results on the generated question-answer pairs providing both positive and negative examples. In Figure~\ref{fig:supp-vqacl-further-qualitatives}-a and Figure~\ref{fig:supp-vqacl-further-qualitatives}-b \OURMODEL{} successfully generates both accurate and contextually appropriate questions and accompanying answers. For instance, questions such as ``What material is the floor made of?'' with the answer ``tile'' and ``Is there a bear in the picture?'' with the answer ``yes'' demonstrate the model's ability to understand and respond correctly based on the visual data. Differently, in Figure~\ref{fig:supp-vqacl-further-qualitatives}-c we highlight a scenario where the generated question is vague and lacks specificity. An example from this row includes a question like ``What color is the jacket?'' which, although correct (answered as ``orange''), does not specify which jacket to look for as there are multiple people wearing jackets in the image. Finally, Figure~\ref{fig:supp-vqacl-further-qualitatives}-d presents instances where the model's generated answers are incorrect. For example, the question ``How many planes are there?'' receives the answer ``3'', whereas we can clearly see there are in fact only 2 airplanes, indicating the model's challenges in some contexts or its misinterpretation of the visual content.

\section{Complexity overhead discussion}
\begin{table}[t!] 
\centering
\caption{Computational overhead in terms of training parameters and time. \texttt{Param count} indicates the fraction of trained parameters, \texttt{TFlops} indicates Tera Flops in forward+backward passes in 1 epoch, \texttt{Computational time} metrics report the processing time requirements (in seconds) per one epoch of a single task in CLOVE-function across different methods.}
\resizebox{\columnwidth}{!}{
\begin{tabular}{lccccc}
\toprule
\multirow{2}{*}{\textbf{Method}} & \multirow{2}{*}{\textbf{Param count}} & \textbf{TFlops} & \multicolumn{3}{c}{\textbf{Computational time (s)}} \\
\cmidrule(lr){4-6}
 & & & \textbf{Training} & \textbf{Generation} & \textbf{Balancing} \\
\midrule
Rehearsal & 0.05 & 2711.05 & 35,085 & n.a. & n.a. \\
LAMOL*    & 0.1  & 5670.19 & 49,650 & 363 & n.a. \\
GaB-clustering (Ours) & 0.3 & 3568.38 & 42,885 & 1,452 & 273 \\
\bottomrule
\end{tabular}%
}
\label{tab:computational-overhead}
\end{table}
As a data-free rehearsal strategy ~\OURMODEL{} trades-off the prohibited access to past tasks data with increased computational complexity for sample generation. More precisely, \OURMODEL{} requires two passes of the employed VLM: a first forward-backward pass trains the qa projection layer $(f_{v\to qa})$ and answer projection layer $(f_{v\to a})$ while the second forward pass allows to generate rehearsal data pairs. Limited overhead is required for balancing the questions according to precomputed real-data clustering statistics, with no need for further modification. Only the task-specific projection heads and answering head are trained, while the rest of the model remains frozen, making the process computationally efficient. We evaluate the computational overhead for pseudo-rehearsal generation and balancing of the presented approach, results are presented in Tab.~\ref{tab:computational-overhead}, where we can see that the approach falls behind the rehearsal baseline while being competitive with the data-free version of LAMOL*. Both \OURMODEL{} and LAMOL* suffer from the longer training time due to the requirements of learning how to generate question-answer pairs, however, \OURMODEL{} avoids replaying for the generation task, ending up with shorter training time at the cost of larger number of training parameters. No additional computation is required at inference time where task-specific heads can be discarded and the single shared answering head solves for the VQA task.

\clearpage
{\small
\bibliographystyle{ieee_fullname}
\bibliography{references}
}